%% file: amr-ner-plus-plus.tex
\documentclass[11pt]{article}
\usepackage{acl2015}
\usepackage{times}
\usepackage{url}
\usepackage{latexsym}
\usepackage{graphicx}
\usepackage{bbm}
\usepackage{mathrsfs}
\usepackage{times,latexsym,amsfonts,amssymb,amsmath,graphicx,url,bbm,rotating,datetime}
\usepackage{enumitem,multirow,hhline,stmaryrd,bussproofs,mathtools,siunitx,arydshln}

\input std-macros.tex

\newcommand\w[1]{\textit{#1}} 
\newcommand\e[1]{\textit{#1}} 
\newcommand\n[1]{\textit{#1}} 

\title{Robust Subgraph Generation Improves Abstract Meaning Representation Parsing}

\author{Keenon Werling \\
  Stanford University \\
  {\tt keenon@stanford.edu} \\\And
  Gabor Angeli \\
  Stanford University \\
  {\tt angeli@stanford.edu} \\\And
  Christopher D. Manning \\
  Stanford University \\
  {\tt manning@stanford.edu} \\}

\date{}

\begin{document}
\maketitle
\begin{abstract}


The Abstract Meaning Representation (AMR) is a representation for open-domain rich semantics, with potential use in fields like event extraction and machine translation. Node generation, typically done using a simple dictionary lookup, is currently an important limiting factor in AMR parsing. We propose a small set of actions that derive AMR subgraphs by transformations on spans of text, which allows for more robust learning of this stage. Our set of construction actions generalize better than the previous approach, and can be learned with a simple classifier. We improve on the previous state-of-the-art result for AMR parsing, boosting end-to-end performance by 3 F$_1$ on both the LDC2013E117 and LDC2014T12 datasets.


\end{abstract}

\section{Introduction}

\Fig{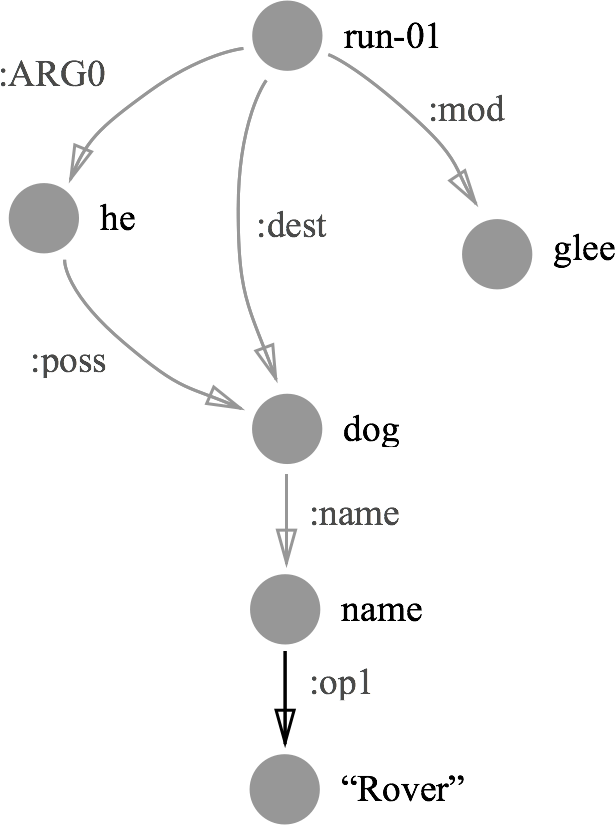}{0.20}{glee}{The AMR graph for 
\w{He gleefully ran to his dog Rover}.
We show that improving the generation of low level subgraphs
  (e.g., \w{Rover} generating \n{name} $\xrightarrow{:op1}$ \n{``Rover''}) significantly improves
  end-to-end performance.
}

The Abstract Meaning Representation (AMR) \cite{2013banarescu-amr} is a rich, graph-based language for expressing semantics over a broad domain.
The formalism is backed by a large data-labeling effort, and it holds
  promise for enabling a new breed of natural language applications ranging from semantically aware MT to rich broad-domain QA over text-based knowledge bases.
\reffig{glee} shows an example AMR for ``he gleefully ran to his dog Rover,'' and we give a brief introduction to AMR in \refsec{crash}.
This paper focuses on AMR parsing, the task of mapping a natural language sentence into an AMR graph.


\FigStar{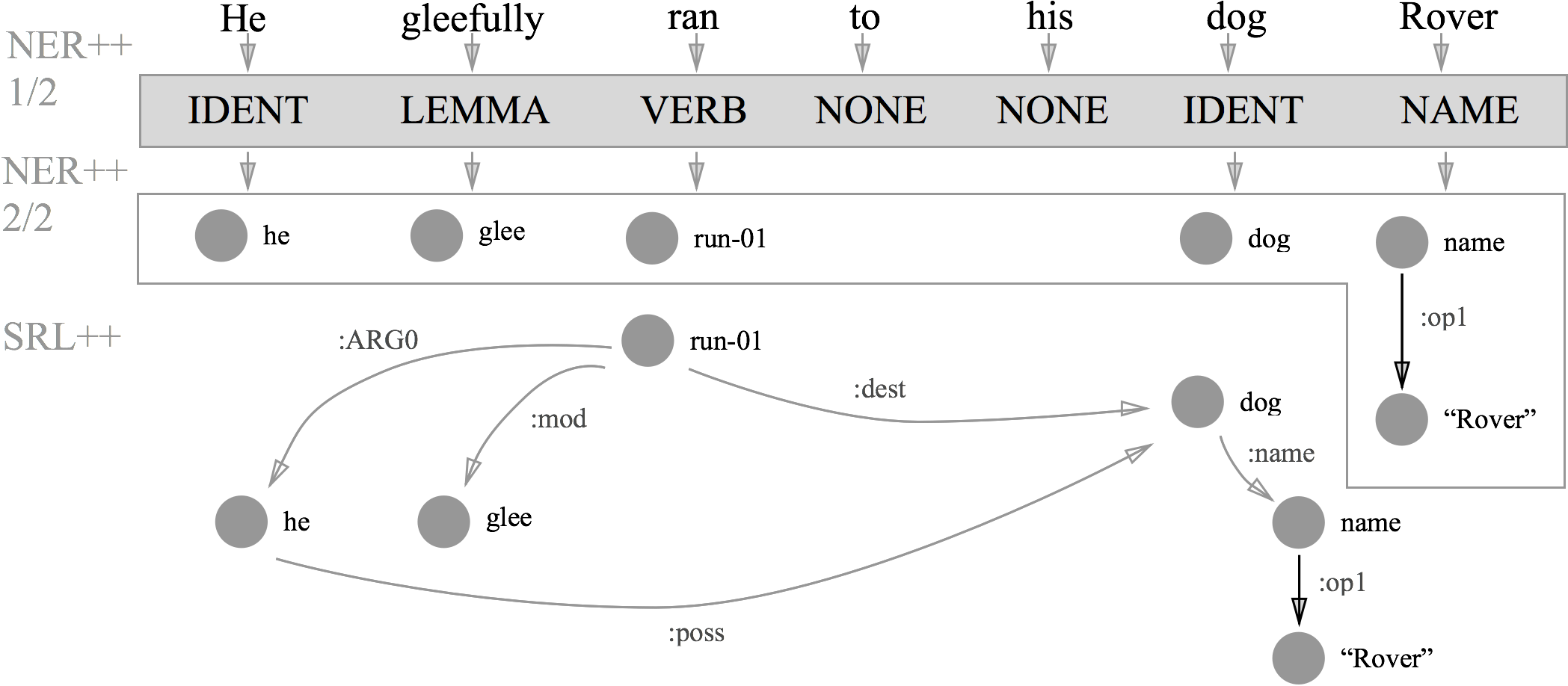}{0.22}{method}{
A graphical explanation of our method. We represent the derivation process for \w{He gleefully ran to his dog Rover}. First the tokens in the sentence are labeled with derivation actions, then these actions are used to generate AMR subgraphs, which are then stitched together to form a coherent whole.}
%
%
%
We follow previous work \cite{2014flanigan-amr} in dividing AMR parsing into two steps. 
The first step is \textit{concept identification}, which generates AMR nodes from text, and which we'll refer to as \textit{NER++} (\refsec{ner}). 
The second step is \textit{relation identification}, which adds arcs to link these nodes into a fully connected AMR graph, which we'll call \textit{SRL++} (\refsec{srl}).

We observe that SRL++ is not the hard part of AMR parsing;
  rather, much of the difficulty in AMR is generating high accuracy concept
  subgraphs from the NER++ component.
For example, when the existing AMR parser JAMR \cite{2014flanigan-amr} is given a gold NER++ output, and must only perform SRL++ 
  over given subgraphs it scores 80 F$_1$ -- nearly the inter-annotator agreement of 83 F$_1$, and far higher than its end to end accuracy of 59 F$_1$.

SRL++ within AMR is \textit{relatively} easy given a perfect NER++ output, because so much pressure is put on the output of NER++ to carry meaningful information.
For example, there's a strong type-check feature for the existence and type of any arc just by looking at its end-points, and syntactic dependency features are very informative for removing any remaining ambiguity.
If a system is considering how to link the node \n{run-01} in \reffig{glee}, the verb-sense frame for ``run-01'' leaves very little uncertainty for what we could assign as an \w{ARG0} arc.
It must be a noun, which leaves either \w{he} or \w{dog}, and this is easily decided in favor of \w{he} by looking for an nsubj arc in the dependency parse.

The primary contribution of this paper is a novel approach to the NER++ task, illustrated in \reffig{method}. We notice that the subgraphs aligned to lexical items can often be generated from a small set of \textit{generative actions} which generalize across tokens. For example, most verbs generate an AMR node corresponding to the verb sense of the appropriate PropBank frame -- e.g., \w{run} generates \n{run-01} in \reffig{glee}. This allows us to frame the NER++ task as the task of classifying one of a small number of \textit{actions} for each token, rather than choosing a specific AMR subgraph for every token in the sentence.


%

Our approach to the end-to-end AMR parsing task is therefore as follows:
  we define an action space for generating AMR concepts, and create a classifier
  for classifying lexical items into one of these actions (\refsec{nerplusplus}).
This classifier is trained from automatically generated alignments between the
  gold AMR trees and their associated sentences (\refsec{alignment}), using an
  objective which favors alignment mistakes which are least harmful to the NER++
  component.
Finally, the concept subgraphs are combined into a coherent AMR parse using the
  maximum spanning connected subgraph algorithm of \newcite{2014flanigan-amr}.

We show that our approach provides a large boost to recall over previous approaches, and that end to end performance is improved from 59 to 62 \n{smatch} (an F$_1$ measure of correct AMR arcs; see \newcite{cai2013smatch-amr}) when incorporated into the SRL++ parser of \newcite{2014flanigan-amr}.
When evaluating the performance of our action classifier in isolation, we obtain
  an action classification accuracy of 84.1\%.

\Section{crash}{The AMR Formalism}

AMR is a language for expressing semantics as a rooted, directed, and potentially cyclic graph, where nodes represent concepts and arcs are relationships between concepts.
AMR is based on neo-Davidsonian semantics, \cite{Davidson:1967,Parsons:1990}.
The nodes (concepts) in an AMR graph do not have to be explicitly grounded in the source sentence, 
and while such an alignment is often generated to train AMR parsers, it is not provided in the training corpora.
The semantics of nodes can represent lexical items (e.g., \w{dog}), sense tagged lexical items (e.g., \textit{run-01}), type markers (e.g., \textit{date-entity}), and a host of other phenomena.

The edges (relationships) in AMR describe one of a number of semantic relationships between concepts.
The most salient of these is semantic role labels, such as the \w{ARG0} and \w{destination} arcs in \reffig{method}.
However, often these arcs define a semantics more akin to syntactic dependencies (e.g., \textit{mod} standing in for adjective and adverbial modification), or take on domain-specific meaning (e.g., the month, day, and year arcs of a \textit{date-entity}).

To introduce AMR and its notation in more detail, we'll unpack the translation of the sentence ``he gleefully ran to his dog Rover.'' 
We show in \reffig{glee} the interpretation of this sentence as an AMR graph.


The root node of the graph is labeled \n{run-01}, corresponding to the PropBank \cite{palmer2005proposition-srl} definition of the verb \w{ran}.
\w{run-01} has an outgoing \e{ARG0} arc to a node \w{he}, with the usual PropBank semantics.
The outgoing \e{mod} edge from \n{run-01} to \n{glee} takes a general purpose semantics corresponding to adjective, adverbial, or other modification of the governor by the dependent.
We note that \n{run-01} has a \e{destination} arc to \n{dog}.
The label for \e{destination} is taken from a finite set of special arc sense tags similar to the preposition senses found in \cite{srikumar2013-srl}.
The last portion of the figure parses \w{dog} to a node which serves as a type marker similar to named entity
types, and \w{Rover} into the larger subgraph indicating a concept with name ``Rover.''

%


\Subsection{sub-chunks}{AMR Subgraphs}
The mapping from tokens of a sentence to AMR nodes is not one-to-one.
A single token or span of tokens can generate a \textit{subgraph} of AMR consisting
  of multiple nodes.
These subgraphs can logically be considered the expression of a single concept,
  and are useful to treat as such (e.g., see \refsec{ner}).

\Fig{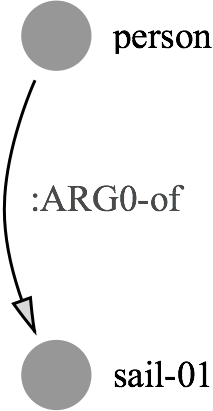}{0.25}{sailor}{AMR representation of the word \w{sailor}, which is notable for breaking the word up into a self-contained multi-node unit unpacking the derivational morphology of the word.}

\Fig{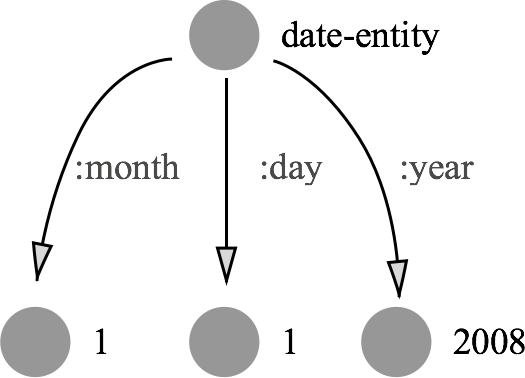}{0.25}{date}{AMR representation of the span \w{January 1, 2008}, an example of how AMR can represent structured data by creating additional nodes such as \n{date-entity} to signify the presence of special structure.}

Many of these multi-node subgraphs capture structured data such as time expressions,
  as in \reffig{date}.
In this example, a \n{date-entity} node is created to signify that this cluster of 
  nodes is part of a structured sub-component representing a date, where the nodes
  and arcs within the component have specific semantics.
This illustrates a broader recurring pattern in AMR: an artificial node may, based on its title, have expected children with special semantics.
A particularly salient example of this pattern is the \n{name} node (see \n{``Rover''} in \reffig{glee}) which signifies 
  that all outgoing arcs with label \e{op} comprise the tokens of a name object.

The ability to decouple the meaning representation of a lexical item from its
  surface form allows for rich semantic interpretations of certain concepts
  in a sentence.
For example, the token \w{sailor} is represented in \reffig{sailor} by a concept graph 
  representing a person who performs the action \n{sail-01}. 
Whereas often the AMR node aligned to a span of text is a straightforward function
  of the text, these cases remain difficult to capture in a principled way beyond
  memorizing mappings between tokens and subgraphs.

\section{Task Decomposition}


To the best of our knowledge, the JAMR parser is
the only published end-to-end AMR parser at the time of publication.
An important insight in JAMR is that AMR parsing can be broken into two 
 distinct tasks: (1) \textbf{NER++} (\textit{concept identification}): the task of interpreting what entities are being referred to in 
the text, realized by generating the best AMR subgraphs for a given set of tokens, and
(2) \textbf{SRL++} (\textit{relation identification}): the task of discovering what 
relationships exist between entities, realized by taking the disjoint subgraphs generated
  by NER++ and creating a fully-connected graph.
We describe both tasks in more detail below.


\Subsection{ner}{NER++}
Much of the difficulty of parsing to AMR lies in generating local subgraphs representing the meaning of token spans.
For instance, the formalism implicitly demands rich notions of NER, lemmatization, word sense disambiguation, number normalization, and temporal parsing; among others.
To illustrate, a correct parse of the sentence in \reffig{method} requires lemmatization (\textit{gleefully} $\rightarrow$ \textit{glee}), word sense tagging (\textit{run} $\rightarrow$ \textit{run-01}), 
and open domain NER (i.e., \textit{Rover}),
Furthermore, many of the generated subgraphs (e.g., \textit{sailor} in \reffig{sailor}) have rich semantics beyond those produced by standard NLP systems.

Formally, NER++ is the task of generating a disjoint set of subgraphs representing the meanings of localized spans of words in the sentence.
For NER++, JAMR uses a simple Viterbi sequence model to directly generate AMR-subgraphs from memorized mappings of text spans to subgraphs.
This paper's main contribution, presented in \refsec{nerplusplus}, is
  to make use of generative \textit{actions} to generate these subgraphs, rather than appealing
  to a memorized mapping.


\Subsection{srl}{SRL++}
The second stage of the AMR decomposition consists of generating a coherent graph
  from the set of disjoint subgraphs produced by NER++.
Whereas NER++ produces subgraphs whose arcs encode domain-specific semantics
  (e.g., \textit{month}), 
  the arcs in SRL++ tend to have generally applicable semantics.
For example, the many arcs encode conventional semantic roles 
  (e.g., \e{ARG0} and \e{destination} in \reffig{method}),
  or a notion akin to syntactic dependencies (e.g., \e{mod} and \e{poss} in \reffig{method}).
For SRL++, JAMR uses a variation of the maximum spanning connected graph algorithm augmented by dual decomposition to impose linguistically motivated constraints on a maximum likelihood objective. 


\section{A Novel NER++ Method}\label{sec:nerplusplus}
The training sets currently available for AMR are not large.
To illustrate, 38\% of the words in the LDC2014E113 dev set are 
  unseen during training time. 
With training sets this small, memorization-based approaches are extremely brittle.
We remove much of the necessity to memorize mappings in NER++
  by partitioning the AMR subgraph search space in terms of the actions needed to 
  derive a node from its aligned token. 
At test time we do a sequence labeling of input tokens with these actions, and 
  then deterministically derive the AMR subgraphs from spans of tokens by applying 
  the transformation decreed by their actions.
We explain in \refsec{actions} how exactly we manage this partition, and in \refsec{data} how we create training data from existing resources to setup and train an action-type classifier.

\Subsection{actions}{Derivation actions}

We partition the AMR subgraph space into a set of 9 actions, each corresponding to an action that will be taken by the NER++ system if a token receives this classification.

\paragraph{IDENTITY} This action handles the common case that the title of the node corresponding to a token is identical to the source token. To execute the action, we take the lowercased version of the token to be the title of the corresponding node.

\paragraph{NONE} This action corresponds to ignoring this token, in the case that
  the node should not align to any corresponding AMR fragment.

\paragraph{VERB} This action captures the verb-sense disambiguation feature of AMR. To execute on a token, we find the most similar verb in PropBank based on Jaro-Winkler distance, and adopt its most frequent sense.
This serves as a reasonable baseline for word sense disambiguation, although of
  course accuracy would be expected to improve if a sophisticated system were
  incorporated.

\paragraph{VALUE} This action interprets a token by its integer value. 
The AMR representation is sensitive to the difference between a node with a title
  of \n{5} (the integer value) and ``5'' or ``five'' -- the string value.
This is a rare action, but is nonetheless distinct from any of the other classes.
We execute this action by extracting an integer value with a regex based number normalizer, and using the result as the title of the generated node.

\paragraph{LEMMA} AMR often performs stemming and part-of-speech transformations on the source token in generating a node. 
For example, we get \n{glee} from \w{gleefully}.
We capture this by a \textbf{LEMMA} action, which is executed by using the lemma of the source token as the generated node title.
Note that this does not capture all lemmatizations, as there are often discrepancies
  between the lemma generated by the lemmatizer and the correct AMR lemma.

\paragraph{NAME} AMR often references names with a special structured data type: the \n{name} construction. 
For example, \w{Rover} in \reffig{glee}.
We can capture this phenomenon on unseen names by attaching a created \n{name} node to the top of a span.

\paragraph{PERSON} A variant of the NAME action, this action produces a subgraph identical to the NAME action, but adds a node \n{person} as a parent. This is, in effect, a \n{name} node with an implicit entity type of person. Due to discrepancies between the output of our named entity tagger and the richer AMR named entity ontology, we only apply this tag to the person named entity tag.

\paragraph{DATE} The most frequent of the structured data type in the data, after \n{name}, is the \n{date-entity} construction (for an example see \reffig{date}).
We deterministically take the output of SUTime \cite{2012chang-temporal}
  and convert it into the \n{date-entity} AMR representation.

\paragraph{DICT} This class serves as a back-off for the other classes, implementing
an approach similar to \newcite{2014flanigan-amr}.
In particular, we memorize a simple mapping from spans of text
  (such as \w{sailor}) to their corresponding most frequently aligned AMR subgraphs 
  in the training data (i.e., the graph in \reffig{sailor}). 
See \refsec{alignment} for details on the alignment process.
At test time we can do a lookup in this dictionary for any element that gets 
  labeled with a \textbf{DICT} action. 
If an entry is not found in the mapping, we back off to the second most probable
  class proposed by the classifier.

It is worth observing at this point that our actions derive much of their power from the similarity 
  between English words and their AMR counterparts; creating an analogue of these actions for other
  languages remains an open problem. 

%
%
\Subsection{informativeness}{Action Reliability}

\Fig{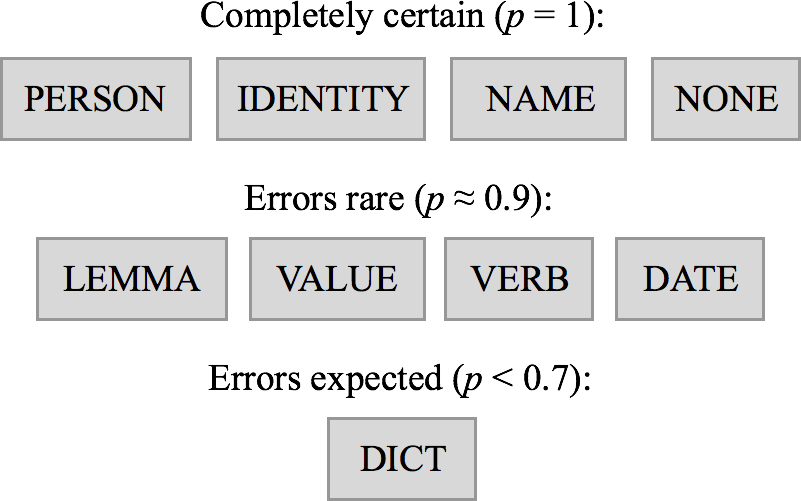}{0.25}{hierarchy}{
Reliability of each action.
The top row are actions which are deterministic;
  the second row occasionally produce errors.
DICT is the least preferred action, with a relatively high error rate.
}

In many cases, multiple actions could yield the same subgraph when applied to a node.
In this section we introduce a method for resolving this ambiguity based on comparing 
the reliability with which actions generate the correct subgraph, and discuss implications.


Even given a perfect action classification for a token, certain action executions can introduce errors.
Some of our actions are entirely deterministic in their conversion
  from the word to the AMR subgraph (e.g., IDENTITY), but others are prone to
  making mistakes in this conversion (e.g., VERB, DICT).
We define the 
  notion of \textit{action reliability} as the probability of deriving the correct node from a span of tokens,
  conditioned on having chosen the correct action.

To provide a concrete example, our dictionary lookup classifier predicts the correct
  AMR subgraph 67\% of the time on the dev set.
We therefore define the reliability of the \textbf{DICT} action as 0.67.
In contrast to \textbf{DICT}, correctly labeling a node as \textbf{IDENTITY}, \textbf{NAME}, \textbf{PERSON}, and \textbf{NONE} have 
  action reliability of 1.0, since there is no ambiguity in the node generation 
  once one of those actions have been selected, 
  and we are guaranteed to generate the correct node given the correct action.

We can therefore construct a hierarchy of reliability (\reffig{hierarchy}) -- all else being equal, we
  prefer to generate actions from higher in the hierarchy, as they are more likely
  to produce the correct subgraph.
This hierarchy is useful in resolving ambiguity throughout our system. 
During the creation of training data for our classifier (\refsec{data}) from our aligner, 
  when two actions could both generate the aligned AMR node we prefer the more reliable one.
In turn, in our aligner we bias alignments towards those which generating 
  more reliable action sequences as training data (see \refsec{alignment}).


The primary benefit of this action-based NER++ approach is that we can reduce the 
usage of low reliability actions, like \textbf{DICT}. 
The approach taken in \newcite{2014flanigan-amr} can be thought of as
equivalent to classifying every token as the \textbf{DICT} action.

%
%

\begin{table}[t]
\begin{center}
\begin{tabular}{l|rr}
\bf Action & \bf \# Tokens & \bf \% Total \\ \hline
NONE & 41538 & 36.2 \\
DICT & 30027 & 26.1 \\
IDENTITY & 19034 & 16.6 \\
VERB & 11739 & 10.2 \\
LEMMA & 5029 & 4.5 \\
NAME & 4537 & 3.9 \\
DATE & 1418 & 1.1 \\
PERSON & 1336 & 1.1 \\
VALUE & 122  & 0.1\\
\end{tabular}
\end{center}
\caption{\label{tab:distro} Distribution of action types in the proxy section of the newswire section of the LDC2014T12 dataset, generated from automatically aligned data. }
\end{table}

We analyze the empirical distribution of actions in our automatically aligned corpus in \reftab{distro}.
The cumulative frequency of the non-\textbf{DICT} actions is striking: we can generate 74\% of the tokens with high reliability ($p \geq 0.9$) actions.
In this light, it is unsurprising that our results demonstrate a large gain in recall on the test set.





%
%
\Subsection{data}{Training the Action Classifier}

\begin{table}[t]
\small
\begin{center}
\begin{tabular}{l}
Input token; word embedding                  \\
Left+right token / bigram                    \\
Token length indicator     \\
Token starts with ``non''     \\
POS; Left+right POS / bigram                 \\
Dependency parent token / POS                \\
Incoming dependency arc                      \\
Bag of outgoing dependency arcs              \\
Number of outgoing dependency arcs           \\
Max Jaro-Winkler to any lemma in PropBank      \\
Output tag of the \textbf{VERB} action if applied \\
Output tag of the \textbf{DICT} action if applied \\
NER; Left+right NER / bigram                 \\
Capitalization                               \\
Incoming prep\_* or appos + parent has NER   \\
Token is pronoun                             \\
Token is part of a coref chain               \\
Token pronoun and part of a coref chain     \\
\end{tabular}
\end{center}
\caption{\label{tab:features} The features for the NER++ maxent classifier. }
\end{table}

Given a set of AMR training data, in the form of \w{(graph, sentence)} pairs,
  we first induce alignments from the graph nodes to the sentence 
  (see \refsec{alignment}).
Formally, for every node $n_i$ in the AMR graph, alignment gives us some token $s_j$ (at the $j$th index in the sentence) that we believe generated the node $n_i$.

Then, for each action type, we can ask whether or not that action type is able to 
  take token $s_j$ and correctly generate $n_i$. 
For concreteness, imagine the token $s_j$ is \w{running}, and the node $n_i$ has 
  the title \n{run-01}.
The two action types we find that are 
  able to correctly generate this node are DICT and VERB. 
We choose the most reliable action type of those available (see \reffig{hierarchy})
  to generate the observed node -- in this case, VERB.

In cases where an AMR subgraph is generated from multiple tokens, we assign the
  action label to each token which generates the subgraph.
Each of these tokens are added to the training set; at test time, we collapse
  sequences of adjacent identical action labels, and apply the action once to the
  resulting token span.
  
Inducing the most reliable action (according to the alignments) for every token in the training corpus provides a supervised training set for our action classifier, with some noise introduced by the automatically generated alignments.
We then train a simple maxent classifier\footnote{A sequence model was tried and showed no improvement over a simple maxent classifier.} to make action decisions at each node.
At test time,
  the classifier takes as input a pair $\langle i, S \rangle$, where $i$ is the 
  index of the token in the input sentence, and $S$ is a sequence tokens 
  representing the source sentence.
It then uses the features in \reftab{features} to predict the actions to take at
  that node.


%
%
%
%

\input alignments.tex

\Section{related}{Related Work}

Prior work in AMR and related formalisms include \newcite{jones2012semantics}, and \newcite{2014flanigan-amr}.
\newcite{jones2012semantics}, motivated by applications in Machine Translation, proposed a graphical semantic meaning representation that predates AMR, but is intimately related.
They propose a hyper-edge replacement grammar (HRG) approach to parsing into and out of this graphical semantic form.
\newcite{2014flanigan-amr} forms the basis of the approach of this paper.
Their system introduces the two-stage approach we use:
  they implement a rule-based alignment to learn a mapping from tokens to subgraphs, and
  train a variant of a maximum spanning tree parser adapted to graphs and 
  with additional constraints for their relation identifications (SRL++) component.
\newcite{wangtransition} uses a transition based algorithm to transform dependency trees into AMR parses.
They achieve 64/62/63 P/R/F$_1$ with contributions roughly orthogonal to our own.
Their transformation action set could be easily augmented by the robust subgraph generation we propose here,
  although we leave this to future work.

Beyond the connection of our work with \newcite{2014flanigan-amr}, we note that 
the NER++ component of AMR encapsulates a number of lexical NLP tasks.
These include named entity recognition \cite{2007nadeau-ner,stanford-ner},
  word sense disambiguation \cite{1995yarowsky-wsd,2002banerjee-wsd},
  lemmatization, and a number of more domain specific tasks.
For example, a full understanding of AMR requires normalizing temporal
  expressions \cite{2010verhagen-tempeval,2010strotgen-temporal,2012chang-temporal}.
 
In turn, the SRL++ facet of AMR takes many insights from semantic role labeling
  \cite{2002gildea-srl,2004punyakanok-srl,srikumar2013-srl,das2014frame} to capture the
  relations between verbs and their arguments.
In addition, many of the arcs in AMR have nearly syntactic interpretations
  (e.g., \e{mod} for adjective/adverb modification, \e{op} for compound noun
  expressions).
These are similar to representations used in syntactic dependency parsing
  \cite{stanford-dependencies,2005mcdonald-dependency0,2006buchholz-conll}.

More generally, parsing to a semantic representation is has been explored in
  depth for when the representation is a logical form
  \cite{2005kate-semantics,2005zettlemoyer-semantics,2011liang-semantics}.
Recent work has applied semantic parsing techniques to representations beyond
  lambda calculus expressions.
For example, work by \newcite{2014berant-bio} parses
  text into a formal representation of a biological process.
\newcite{2014hosseini-algebra} solves algebraic word problems by parsing them
  into a structured meaning representation.
In contrast to these approaches, AMR attempts to capture open domain semantics
  over arbitrary text.

Interlingua
  \cite{1991mitamura-interlingua,1999carbonell-interlingua,1998levin-interlingua}
  are an important inspiration for decoupling the semantics of the AMR language
  from the surface form of the text being parsed; although, AMR has a self-admitted
  English bias.

\section{Results}
We present improvements in end-to-end AMR parsing on two datasets  using our NER++ component.
Action type classifier accuracy on an automatically aligned corpus
and alignment accuracy on a small hand-labeled corpus are also reported.

\Subsection{result-amr}{End-to-end AMR Parsing}
We evaluate our NER++ component in the context of end-to-end AMR parsing
on two corpora: the newswire section of LDC2014T12 and the split given in \newcite{2014flanigan-amr} of LDC2013E117, both consisting primarily of newswire.
We compare two systems: the JAMR parser \cite{2014flanigan-amr},\footnote{Available at \url{https://github.com/jflanigan/jamr}.}
  and the JAMR SRL++ component with our NER++ approach.

AMR parsing accuracy is measured with a metric called \w{smatch} \cite{cai2013smatch-amr}, which stands 
  for ``s(emantic) match.'' 
The metric is the F$_1$ of a best-match between triples implied by the target graph, 
  and triples in the parsed graph -- that is, the set of (parent, edge, child) triples
  in the graph.

Our results are given in \reftab{results}.
We report much higher recall numbers on both datasets, with only small ($\leq 1$ point) 
  loss in precision.
This is natural considering our approach.
A better NER++ system allows for more correct AMR subgraphs to be generated --
  improving recall -- but does not in itself necessarily improve the accuracy of the
  SRL++ system it is integrated in.


\begin{table}[t]
\begin{center}
\begin{tabular}{l|l|llr}
\textbf{Dataset} &  \textbf{System} & \textbf{P} & \textbf{R} & \textbf{F$_1$} \\
\hline
\multirow{2}{*}{2014T12} & JAMR & \textbf{67.1} & 53.2 & 59.3 \\
  & \textbf{Our System} & 66.6 & \textbf{58.3} & \textbf{62.2} \\
\hline
\multirow{2}{*}{2013E117} & JAMR & \textbf{66.9} & 52.9 & 59.1 \\
  & \textbf{Our System} & 65.9 & \textbf{59.0} & \textbf{62.3} \\
\end{tabular}
\end{center}
\caption{\label{tab:results} 
Results on two AMR datasets for JAMR and our NER++ embedded in the JAMR SRL++
  component.
Note that recall is consistently higher across both datasets, with only a small
  loss in precision.
}
\end{table}

\Subsection{result-component}{Component Accuracy}
We evaluate our aligner on a small set of 100 hand-labeled alignments,
  and evaluate our NER++ classifier on automatically generated alignments over the whole corpus,
  
On a hand-annotated dataset of 100 AMR parses from the LDC2014T12 corpus,\footnote{
    Our dataset is publicly available at \url{http://nlp.stanford.edu/projects/amr}
  }
  our aligner achieves
  an accuracy of \textbf{83.2}.
This is a measurement of the percentage of AMR nodes that are
  aligned to the correct token in their source sentence.
Note that this is a different metric than the precision/recall of prior work on alignments, and
  is based on both a different alignment dataset and subtly different alignment annotation scheme.
In particular, we require that every AMR node aligns to some token in the sentence,
  which forces the system to always align nodes, even when unsure.
A standard semantics and annotation guideline for AMR alignment is left for 
  future work; our accuracy should be considered only an informal metric.

We find our informativeness-based alignment objective slightly improves end-to-end performance
  when compared to the rule-based approach of \cite{2014flanigan-amr}, 
  improving F$_1$ by roughly 1 point (64/59/61 P/R/F$_1$ to 65/59/62 P/R/F$_1$).


On the automatic alignments over the LDC2014T12 corpus,
  our action classifier achieved a test accuracy of \textbf{0.841}.
The classifier's most common class of mistakes are incorrect \textbf{DICT} classifications. 
It is reassuring that some of these errors can be recovered from by the 
  \naive\ dictionary lookup finding the correct mapping.

The \textbf{DICT} action lookup table achieved an accuracy of \textbf{0.67}.
This is particularly impressive given that our model moves many of the difficult 
  semantic tasks onto the \textbf{DICT} tag, and that this lookup does not make
  use of any learning beyond a simple count of observed span to subgraph mappings.


\Section{conclude}{Conclusion}
We address a key challenge in AMR parsing: 
  the task of generating subgraphs from lexical items in the sentence.
We show that a 
  simple classifier over \textit{actions} which generate
  these subgraphs improves end-to-end recall for AMR parsing with only a small
  drop in precision, leading to an overall gain in F$_1$.
A clear direction of future work is improving the coverage of the defined actions.
For example, a richer lemmatizer could shift the burden of lemmatizing unknown
  words into the AMR lemma semantics and away from the dictionary lookup component.
We hope our decomposition provides a useful framework to guide future work in NER++ and AMR in general.
  

%


\section*{Acknowledgments}

We thank the anonymous reviewers for their thoughtful feedback.
Stanford University gratefully acknowledges the support of the Defense
Advanced Research Projects Agency (DARPA) Deep Exploration and Filtering
of Text (DEFT) Program under Air Force Research Laboratory (AFRL)
contract no. FA8750-13-2-0040. Any opinions, findings, and conclusion or
recommendations expressed in this material are those of the authors and
do not necessarily reflect the view of the DARPA, AFRL, or the US
government.

\bibliographystyle{acl}
\bibliography{ref}

\end{document}

%% file: std-macros.tex
\newcommand\sE{\ensuremath{\mathcal{E}}}

\newcommand\sT{\ensuremath{\mathcal{T}}}

\newcommand\bN{\ensuremath{\mathbf{N}}}

\newcommand\bQ{\ensuremath{\mathbf{Q}}}

\newcommand\bS{\ensuremath{\mathbf{S}}}

\newcommand\bV{\ensuremath{\mathbf{V}}}

\newcommand\Fig[4]{\begin{figure}[tb] \begin{center} \includegraphics[scale=#2]{#1} \end{center} \caption{\label{fig:#3} #4} \end{figure}}

\newcommand\FigStar[4]{\begin{figure*}[tb] \begin{center} \includegraphics[scale=#2]{#1} \end{center} \caption{\label{fig:#3} #4} \end{figure*}}

\newcommand\refeqn[1]{(\ref{eqn:#1})}

\newcommand\refsec[1]{Section~\ref{sec:#1}}

\newcommand\reffig[1]{Figure~\ref{fig:#1}}

\newcommand\reftab[1]{Table~\ref{tab:#1}}

\newcommand\Section[2]{\section{#2}\label{sec:#1}}
\newcommand\Subsection[2]{\subsection{#2}\label{sec:#1}}

\newcommand\naive{na\"{\i}ve}

\usepackage{color}

%% file: alignments.tex
\Section{alignment}{Automatic Alignment of Training Data}
AMR training data is in the form of bi-text, where we are given a set of
  (sentence, graph) pairs, with no explicit alignments between them.
We would like to induce a mapping from each node in the AMR graph to the token it represents. 
It is perfectly possible for multiple nodes to align to the same token -- this is the case
  with \textit{sailors}, for instance. 

It is not possible, within our framework, to represent a single node being sourced from multiple tokens.
Note that a subgraph can consist of many individual nodes; in cases where a subgraph
  should align to multiple tokens, we generate an alignment from the subgraph's nodes
  to the associated tokens in the sentence.
It is empirically very rare for a subgraph to have more nodes than the token span
  it should align to.




There have been two previous attempts at producing automatic AMR alignments. 
The first was published as a component of JAMR, 
  and used a rule-based approach to perform alignments.
This was shown to work well on the sample of 100 hand-labeled sentences used to 
  develop the system.
\newcite{2014pourdamghani-amr} approached the alignment problem in 
  the framework of the IBM alignment models.
They rendered AMR graphs as text, and then used traditional machine translation 
  alignment techniques to generate an alignment.

We propose a novel alignment method, since
  our decomposition of the AMR node generation process into a set of actions 
  provides an additional objective for the aligner to optimize, in addition to the
  accuracy of the alignment itself.
We would like to produce the most \textit{reliable}
  sequence of actions for the NER++ model to train from, where reliable is taken
  in the sense defined in \refsec{informativeness}.
To give an example, a sequence of all \textbf{DICT} actions could generate any
  AMR graph, but is very low reliability.
A sequence of all \textbf{IDENTITY} actions could only generate one set of nodes, but does it with absolute certainty.




We formulate this objective as a Boolean LP problem.
Let $\bQ$ be a matrix in $\{0,1\}^{|\bN| \times |\bS|}$ of Boolean constrained variables,
  where $\bN$ are the 
  nodes in an AMR graph, and $\bS$ are the tokens in the sentence.
The meaning of $\bQ_{i,j} = \mathbbm{1}$ can be interpreted as node 
  $n_i$ having being aligned to token $s_j$.
Furthermore, let $\bV$ be a matrix $\sT^{|\bN| \times |\bS|}$, where
  $\sT$ is the set of NER++ actions from \refsec{nerplusplus}.
Each matrix element $\bV_{i,j}$ is assigned the most reliable action which would
  generate node $n_i$ from token $s_j$.
We would like to maximize the probability of the actions collectively generating a perfect set of nodes.
This can be formulated linearly by maximizing the log-likelihood of the actions.
Let the function $\textrm{\small{REL}}(l)$ be the reliability of action $l$ (probability of generating intended node).
Our objective can then be formulated as follows:
\begin{align}
  \label{eqn:objective}
  & \underset{\bQ}{\textrm{max}}
     & & \sum_{i,j} \bQ_{i,j} \left[ 
         \log(\textrm{\small{REL}}(\bV_{i,j}))
         + \alpha \sE_{i,j} \right] \\
  \label{eqn:constraint1}
  & \textrm{s.t.}
     & & \sum_{j} \bQ_{i,j} = 1 ~~~~~ \forall i \\
  \label{eqn:constraint2}
  & & & \bQ_{k,j} + \bQ_{l,j} \leq 1 ~~~~~ \forall k,l,j; ~ n_k \nleftrightarrow n_l
\end{align}
where $\sE$ is the Jaro-Winkler similarity between the title of the node $i$ and the
  token $j$, $\alpha$ is a hyperparameter (set to 0.8 in our experiments),
  and the operator $\nleftrightarrow$ denotes that two nodes in the AMR graph are
  both not adjacent and do not have the same title.

The constraint \refeqn{constraint1}, combined with the binary constraint on $\bQ$, ensures that every node in the graph is
  aligned to exactly one token in the source sentence.
The constraint \refeqn{constraint2} ensures that only adjacent nodes or nodes that share a title can refer to the same token.

The objective value penalizes alignments which map to the unreliable DICT tag,
  while rewarding alignments with high overlap between the title of the node and
  the token.
Note that most incorrect alignments fall into the DICT class by default, as no other
  action could generate the correct AMR subgraph.
Therefore, if there exists an alignment that would consume the token using another
  action, the optimization prefers that alignment.
The Jaro-Winkler similarity term, in turn, serves as a tie-breaker between equally
  (un)reliable alignments.

%
%
%
%
%

There are many packages which can solve this Boolean LP efficiently.
We used Gurobi \cite{gurobi}.
Given a matrix $\bQ$ that maximizes our objective, we can decode our solved alignment 
  as follows: for each $i$, align $n_i$ to the $j$ s.t. $\bQ_{i,j} = 1$. 
By our constraints, exactly one such $j$ must exist.